\pdfoutput=1

\documentclass[11pt]{article}

\usepackage[final]{acl}

\usepackage{times}
\usepackage{latexsym}
\usepackage{amsmath}
\usepackage{multirow}
\usepackage{graphicx}
\usepackage{subcaption}
\usepackage{caption}
\usepackage[T1]{fontenc}

\usepackage[utf8]{inputenc}

\usepackage{microtype}

\usepackage{inconsolata}
\usepackage{booktabs}

\usepackage{graphicx}

\usepackage{todonotes}

\definecolor{ocr}{HTML}{00C8FF}
\definecolor{ocr}{HTML}{009900}
\definecolor{joeColor}{rgb}{0.6, 0.2, 1.0}
\definecolor{jackColor}{rgb}{1.0, 0.6, 0.4}
\definecolor{nafisColor}{rgb}{0.5, 0.2, 0.8}
\definecolor{solhaColor}{rgb}{0.2, 0.6, 0.9}

\title{Catch Me If You Can? Not Yet: LLMs Still Struggle to Imitate \\the Implicit Writing Styles of Everyday Authors}

\author{
Zhengxiang Wang\textsuperscript{1}\thanks{Equal contributions.} \hspace{.1cm} 
Nafis Irtiza Tripto\textsuperscript{2}\footnotemark[1] \hspace{.1cm} 
Solha Park\textsuperscript{1} \hspace{.1cm} 
Zhenzhen Li\textsuperscript{3} \hspace{.1cm} 
Jiawei Zhou\textsuperscript{1} \\
\textsuperscript{1}Stony Brook University \hspace{.1cm} 
\textsuperscript{2}The Pennsylvania State University \hspace{.1cm} 
\textsuperscript{3}Bosch Center for AI \\ 
\texttt{zhengxiang.wang@stonybrook.edu, nit5154@psu.edu}
}

\begin{document}
\maketitle
\begin{abstract}

As large language models (LLMs) become increasingly integrated into personal writing tools, a critical question arises: \textit{can LLMs faithfully imitate an individual’s writing style from just a few examples?} Personal style is often subtle and implicit, making it difficult to specify through prompts yet essential for user-aligned generation.
This work presents a comprehensive evaluation of state-of-the-art LLMs’ ability to mimic personal writing styles via in-context learning from a small number of user-authored samples. We introduce an ensemble of complementary metrics—including authorship attribution, authorship verification, style matching, and AI detection—to robustly assess style imitation. Our evaluation spans over 40,000 generations per model across domains such as news, email, forums, and blogs, covering writing samples from more than 400 real-world authors.
Results show that while LLMs can approximate user styles in structured formats like news and email, they struggle with nuanced, informal writing in blogs and forums. Further analysis on various prompting strategies such as number of demonstrations reveal key limitations in effective personalization. Our findings highlight a fundamental gap in personalized LLM adaptation and the need for improved techniques to support implicit, style-consistent generation. To aid future research and for reproducibility, we open-source our data and code.\footnote{\url{https://github.com/jaaack-wang/llms-implicit-writing-styles-imitation}.}

\end{abstract}

\section{Introduction}

Every time humans write, they leave behind a unique linguistic fingerprint \cite{svartvik1968evans_linguistic_fingerprint}, a subtle and subconscious reflection of their personal style. To enhance their writing, humans have been using writing tools from spell and grammar checkers to paraphrasers for quite some time \cite{ferris2004grammar}. But, the rise of large language models (LLMs) has introduced a powerful, all-in-one assistant capable of drafting, editing, and rephrasing text \cite{wasi2024llms}. Despite their impressive fluency, LLMs often default to a \textit{generic} style learned from vast web data, stripping away the personal touch that makes writing feel authentic \cite{padmakumar2023does}. This raises concerns about authorship dilution and being flagged as AI-generated content \cite{tripto2024ship}.

\begin{figure}[t]
    \centering
    \includegraphics[width=\linewidth]{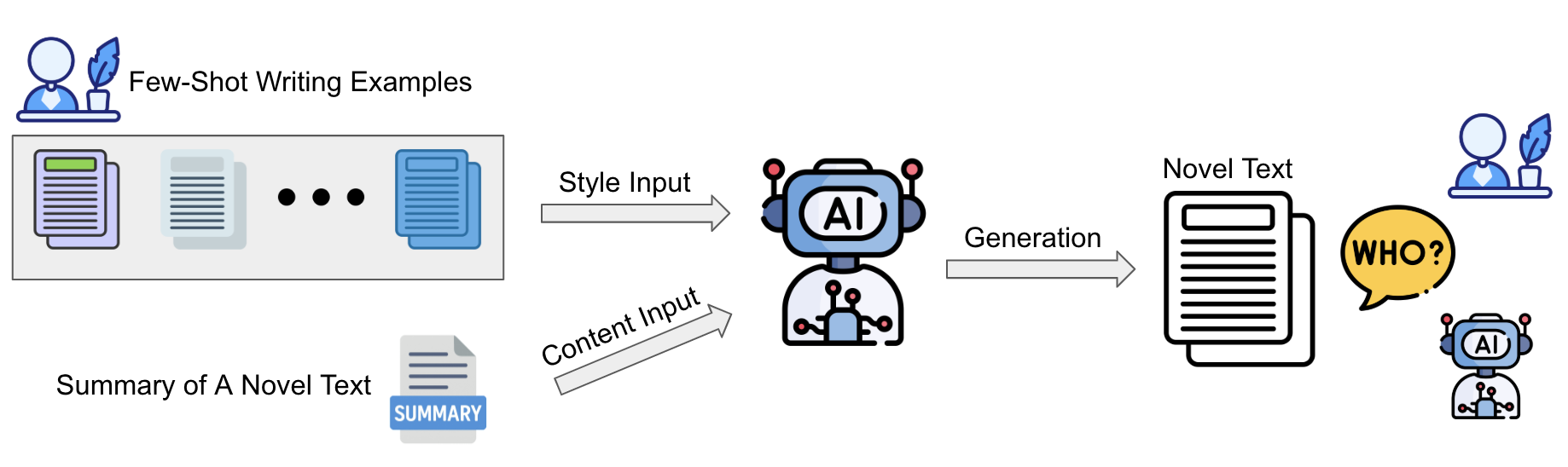}
    \caption{Task of implicit personalized writing imitation: given few-shot writing examples from an everyday author and a content summary, an LLM is prompted to generate a novel text that imitates the writing examples in styles and reflect the semantics of the summary.}
    \label{fig:styleMimickingIllustration}
\end{figure}

To counter these concerns, some systems (e.g., Claude AI\footnote{\url{https://claude.ai/}, accessed May 2025.}) offer configurable parameters like tone, voice, and formality. Yet these controls fall short as users' personal styles are nuanced and rarely reducible to a few sliders \cite{kang2021style}. A more natural solution is to guide the model with examples of an individual's prior writing, employing few-shot prompting to emulate an individual’s style more faithfully, like the task shown in Figure~\ref{fig:styleMimickingIllustration}. While LLMs have shown promise in mimicking public figures \cite{herbold2024large, chen2024using} and fictional characters \cite{li2023chatharuhi} whose extensive data footprints on the internet make them easy to model, their ability to replicate the style of everyday users remains largely unexplored. 
\textit{Can LLMs truly adapt to the personal style of an everyday author with only a handful of casual interactions and no explicitly defined stylistic identities?}

Existing research on personal-style generation for ordinary users is limited \cite{bhandarkar2024emulating, cho-etal-2025-tuning}, and even more pressing is the lack of robust evaluation: how do we determine if a generated text from an LLM can belong to the same individual in style and voice? While the concept of authorship is inherently subjective,computational methods like Authorship Attribution (AA) and Authorship Verification (AV)—well-established in forensic linguistics—offer promising proxies. The theory of Linguistic Individuality \cite{nini2023theory} suggests that a person’s writing style often forms a consistent style model. If LLMs can generate text that fits within this style model while being recognized as from the same author by the two computational authorship models, it brings us closer to achieving truly personalized text generation.

In this work, we examine whether LLMs can imitate the \textit{implicit} writing styles of everyday users through in-context learning, without access to explicit stylistic instructions, via the generation task shown in Figure~\ref{fig:styleMimickingIllustration}. Our investigation is motivated by and rooted in a \textit{realistic} usage scenario: most users interact with LLMs via standard interfaces rather than custom plug-ins or fine-tuned pipelines \cite{wang2024understanding}. While methods like personalized fine-tuning \citep{liu2024customizing} or advanced prompting strategies \citep{cho-etal-2025-tuning} can improve stylistic alignment, they are often impractical in time-sensitive, real-world settings. Instead, we focus on evaluating the \textit{out-of-the-box} ability of current LLMs to replicate personal writing style by conditioning on a few user-authored samples, reflecting the kind of implicit personalization that is most accessible and scalable in practice.

To this end, we conduct a comprehensive evaluation of state-of-the-art LLMs on the task of \textit{personalized writing imitation}, defined as emulating an individual’s implicit writing style solely from prior writing samples, without explicit stylistic instructions and semantically conditioned on a content summary (see Figure~\ref{fig:styleMimickingIllustration}). Given the inherent difficulty of measuring personal style, we adopt a diverse suite of computational evaluators, including authorship attribution, authorship verification, linguistically grounded style metrics, and state-of-the-art AI detection tool. This ensemble approach enables a robust and multifaceted assessment of stylistic fidelity and human-like generation.

Using this framework, we evaluate frontier models from leading providers, including OpenAI, Google, Meta, and DeepSeek, across writing domains such as news articles, emails, online forums, and personal blogs. These domains reflect a range of everyday writing scenarios with distinct stylistic constraints. Our datasets span over 400 authors and build on prior work to ensure diversity and realism. We investigate whether few-shot demonstrations of a user’s prior writing are sufficient for LLMs to generate outputs that match the author's voice and style under our evaluators.

Results show that while LLMs can partially emulate user style in more structured formats like news and email, they struggle with nuanced, informal expression in domains such as blogs and forums. Generated outputs often default to an average, generic tone and remain readily detectable as AI-written. Moreover, increasing the number of demonstrations offers limited gains in stylistic alignment. These findings underscore the current limitations of in-context personalization and highlight the need for more effective methods to achieve truly personalized generation, an essential step toward democratizing LLM-based writing tools.

\section{Related Work}

Personalization in LLMs spans a wide spectrum, from persona emulation \citep{tseng2024two} to personalized content recommendations \citep{lyu2024llm_recommendation}. In this study, however, we specifically focus on the subtle challenge of \textit{implicit personalized writing  imitation}, whether LLMs can generate text that reflects the inherent implicit style learned from an individual’s prior writings. To situate our work, we review progress at the intersection of LLM-based personalized writing and authorship analysis around LLM generated texts.

\begin{figure*}[htbp]
    \centering
    \includegraphics[width=0.9\linewidth]{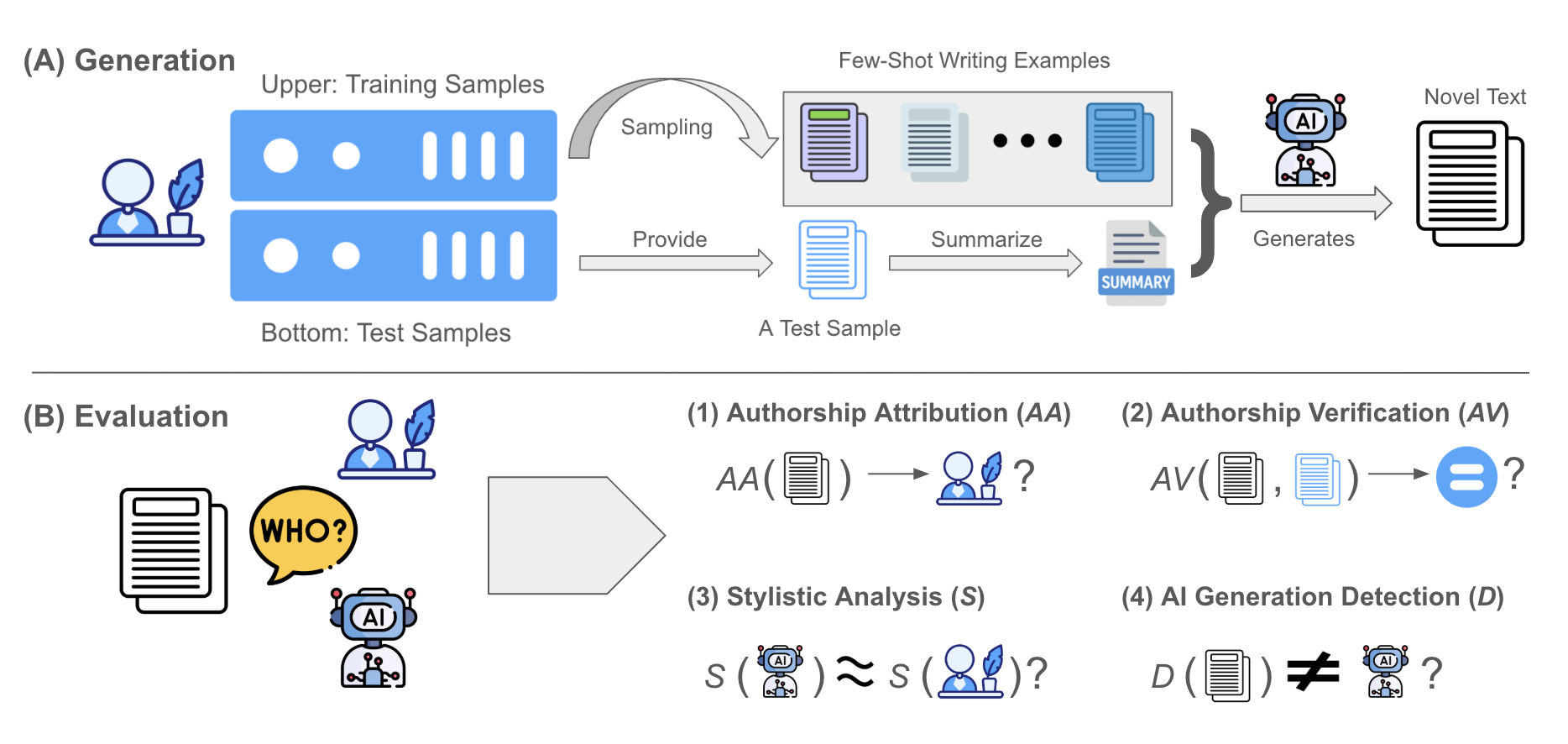}
    \caption{Methodology of our study. (A) We partition each author's writing samples into training and test sets. We select few-shot examples from training samples to guide an LLM to generate novel text based on a test sample summary. (B) We evaluate the extent to which the LLM imitates an author's style using four methods: (1) \textbf{AA}: Does the model predict LLM-generated samples as written by the target author? (2) \textbf{AV}: Does the model predict LLM-generated and human-authored samples as coming from the same author? (3) \textbf{Stylistic Analysis}: How similar are the stylistic distributions of LLM-generated and human-authored samples? (4) \textbf{AI Generation Detection}: Can LLM-generated samples imitating human writing be detected as non-AI texts?}
    \label{fig:methodologyIllustration}
\end{figure*}

\paragraph{LLMs for Personalized Writings}
Recent advances in LLMs have enabled impressive performance in general text style transfer \cite{toshevska2025llm}, with a growing interest in personalized text generation \cite{zhang2024personalization}. Much of prior work focuses on replicating the style of well-known figures, such as authors \cite{mikros2025beyond,dinu2025analyzing}, celebrities \cite{herbold2024large,chen2024using}, or fictional characters \cite{li2023chatharuhi,park2025charactergpt}, where models benefit from ample training data and rich backstories. However, research on replicating the style of ordinary individuals remains underexplored, especially in terms of a rigorous and systematic evaluation.

Some recent studies have attempted individualized generation using retrieval-augmented generation (RAG) systems \cite{mysore2024pearl} or advanced in-context prompting strategies like Trial-Error-Explain \cite{cho-etal-2025-tuning}. Yet, these approaches are often costly, impractical for real-time personalization, or dependent on repeated LLM queries. Crucially, these studies usually rely on an LLM-as-a-judge approach \cite{zheng2023judgingllmasajudge} to measure stylistic alignment. While LLMs have shown remarkable performance in authorship tasks \cite{huang2024can}, their judgment is still subjective and might contain bias for their generation \cite{panickssery2024llm}.

Other evaluations have used metrics like ROUGE or METEOR \cite{kumar2024longlamp}, which assess content overlap but overlook stylistic fidelity. Authorship Attribution (AA) has been used \cite{bhandarkar2024emulating},  but typically in narrow domains or limited to single-text completions without broader style considerations. Therefore, our study introduces a comprehensive evaluation framework grounded in authorship and stylometric analysis, aiming to systematically assess whether LLMs can truly replicate an individual's writing style

\paragraph{Authorship Analysis Tasks}
To evaluate whether LLMs can truly replicate individual writing styles, we draw from established authorship analysis tasks. Authorship Attribution (AA) identifies the author of a given text from a set of candidate authors \cite{kjell1994discrimination}, while Authorship Verification (AV) determines whether two texts share the same author \cite{halteren2007author}. These tasks traditionally rely on features such as n-grams and stylometric cues, e.g., lexical, syntactic, semantic, and structural patterns \citep{stamatatos2009survey, koppel2012fundamental,juola2013overview,  neal2017surveying}. While classic ML and DL models have long been used for stylometry, fine-tuned transformer models now represent state-of-the-art AA and AV tasks \cite{Tyo_Dhingra_Lipton_2022_AV, tripto2023hansen}. Recently, stylometric analysis has also been employed to detect LLM-generated text \cite{herbold2023large, munoz2023contrasting, casal2023can}, offering a promising path to evaluate whether LLM outputs align with a target individual's writing style.

\begin{table*}[]
    \centering
    \resizebox{\textwidth}{!}{
    \begin{tabular}{lllllllll}
    \toprule
    Dataset & Genre & \# Authors & \# Samples & Avg Length & AV Acc. (\%) & AA Top-5 Acc. (\%) & Style Acc. (\%) & \% Human \\ \midrule

    Enron & Emails & 150 & 3,884 & 309 & 88.9 & 79.8 & 79.4 & 99.2 \\
    Blog & Blogs & 100 & 25,224 & 319 & 91.4 & 95.5 & 81.9 & 99.5 \\

    CCAT50 & News articles & 50 & 2,500 & 584 & 89.2 & 94.9 & 69.1 & 100\\

    Reddit & Online forums & 100 & 8,451 & 333 & 87.7 & 89.7 & 73.8 & 100 \\
    
    \bottomrule
    \end{tabular}
    } %
    \caption{Four test sets used in our study and their original dataset sources. The corresponding train sets has same ``\# Authors'' and ``\# Samples'' and similar ``Avg Length.'' The ``AV Acc.'' and ``AA Top-5 Acc.'' refer to the average test set accuracy of the corresponding AV and AA models trained on the related train sets. ``\% Style Acc.'' compute the percentage of times test samples of an author closer to that particular author’s style model compared to other authors’ style model. ``\% Human'' is the percentage of times test set samples detected as human generated.}
 
    \label{tab:datasets}
\end{table*}

\section{Methodology\label{sec:methodology}}

This study evaluates whether LLMs can mimic a user's implicit writing style using only a few prior examples without any explicit stylistic instructions. We consider a realistic scenario, as visualized in Figure~\ref{fig:styleMimickingIllustration}, where users provide a handful of writing samples and a content summary, reflecting typical few-shot prompting, without describing their style in detail (which may also be challenging if feasible). This setup requires no model customization and offers a scalable path to personalization. While writing styles can evolve, a phenomenon studied in stylochronometry \citep{can2004change}, they primarily focus on long-term changes in the writing of novelists or public figures \citep{stamou2007stylochronometry, klaussner2015stylochronometry}. In our context, we assume that individuals tend to maintain a relatively stable style in everyday writing, making this assumption both practical and justified.

Figure~\ref{fig:methodologyIllustration} provides an overall visual illustration of our methodology with the rest of the section structured as follows. Section \ref{subsec_task_setup} defines the task setup, \ref{sec:evaluation} presents the evaluation framework, \ref{subsec_dataset} describes the datasets, and
\ref{subsec_task_rationales} discusses our design choices and limitations.

\subsection{Task Setup and Notations}
\label{subsec_task_setup}

We formalize the task of implicit personalized writing imitation  and describe the task setup and notations. Given a human author $a \in A$, where $A$ is the set of all authors, let $T_a$ denote the collection of distinct writing samples available from author $a$. We split $T_a$ into a training set $T_a^{tr}$ for sampling few-shot writing examples and a test set $T_a^{te}$ for evaluation. For any sample $t_a \in T_a^{te}$, we prompt an LLM $L$ with $n$ writing samples $T_a^{tr'} \subseteq T_a^{tr}$ and a content summary $S_{t_a}$ of $t_a$ to generate a new text $t_a^L$, or formally, $L(T_a^{tr'}, S_{t_a}) \rightarrow t_a^L$. See Figure~\ref{fig:methodologyIllustration} (A) for an illustration. Here, the summary $S_{t_a}$ is used to fix the content (or semantics) of the generated text, so that we can focus our evaluation on how well $t_a^L$ aligns with the author's writing style in a semantically controlled condition. 

We evaluate the implicit personalized writing imitation capabilities of an LLM $L$ by assessing both the stylistic fidelity of the generated text $t_a^L$ to the overall writing style of author $a$ and how closely $t_a^L$ aligns with the specific test sample $t_a$ in terms of inherent style.

\subsection{Personalized Writing Evaluation Framework}
\label{sec:evaluation}

To assess the extent to which LLMs can replicate implicit personal writing style, we propose a multi-angle evaluation framework grounded in computational authorship analysis and human-likeness detection, illustrated in Figure~\ref{fig:methodologyIllustration} (B). This framework contains four components as follows.

\begin{itemize}
    \item An \textbf{authorship attribution model} $\text{AA}(t) \rightarrow a$, which predicts the most likely author $a \in A$ for a given text $t$. It is trained on human-authored texts (training samples from authors, $T_a^{tr}$) and used to test whether the generated text is correctly attributed to the target author.
    
    \item An \textbf{authorship verification model} $\text{AV}(t, t') \rightarrow \{0, 1\}$, which returns 1 if both texts $t$ and $t'$ are judged to be written by the same author, and 0 otherwise. This model is applied to compare each generation with its corresponding human-written reference text.

    \item A \textbf{style model} $X_a$ for each author $a \in A$, built from the distributional and stylistic features of their writing samples $T_a$. We then compute the stylistic distance between a generated text $t$ and each author’s style model $X_a$ to assess stylistic proximity. In contrast to AA and AV models, which capture \textit{implicit} style patterns through fine-tuning, this approach provides an explicit criterion for evaluating whether the generated text mimics an individual’s style.

   \item An \textbf{AI generation detector} $D(t)$ that classifies whether a text $t$ appears human-written or AI-generated. We use GPTZero\footnote{\url{https://gptzero.me/}.}
 for its strong performance in AI text detection. This complements the other three metrics by assessing human-likeness, providing an additional perspective beyond stylistic fidelity.

\end{itemize}

Together, these metrics allow us to computationally probe whether LLM-generated text exhibits stylistic consistency with the target author, enabling reliable evaluation of personalized generation.

\subsection{Data Domains}
\label{subsec_dataset}

We use portions of the four datasets listed in Table~\ref{tab:datasets} for our evaluation: Enron \cite{klimt2004enron}, Blog \cite{schler2006blog_corpus}, CCAT50 \cite{liu2011reuter}, Reddit \cite{volske-etal-2017-reddit_dataset}. These four datasets present four wildly different genres (emails, blogs, news articles, and online forums) and contain a wide range of average authors (400 in total). In terms of formats, while emails and news articles are more structured and formal, blogs and forums are more casual and informal. These features of the included datasets offer a comprehensive testbed for evaluating LLM's stylistic mimicry capabilities across genres in a real-world setting. 

We perform careful sampling with length control on the original datasets and evenly split the selected samples into train and test sets (See Appendix \ref{appendix_data_split} for details). Besides length control, our sampling is randomized. This helps ensure the representativeness of the obtained samples, considering the large scale of the sampling.

\subsection{Discussion of Task Design Rationale}
\label{subsec_task_rationales}
While fine-tuning is often viewed as a reliable method for achieving personalized generation \cite{tan2024democratizing, zhang2024personalization}, we intentionally avoid it in our study. First, real-world users typically provide only a limited number of writing samples, which is insufficient for effective fine-tuning. Second, maintaining separate fine-tuned models for each user is costly and impractical at scale \cite{han2024parameter}. 

Finally, complex prompting techniques that involve multiple back-and-forth iterations with the LLMs, such as \citet{cho-etal-2025-tuning} and \citet{bhandarkar2024emulating}, would introduce significant latency and API overhead, making them unsuitable for real-time use \cite{shekhar2024towards}.

Instead, we adopt a more practical approach grounded in how users typically interact with LLMs via a graphical interface, using only a few prior writing samples stored from earlier interactions. We focus on few-shot prompting, requiring just a single API call, and compare with the zero-shot scenario, used as a baseline to contextualize the effectiveness of few-shot prompting. As emphasized throughout this paper, our primary goal is to rigorously evaluate how well LLMs can imitate an individual's writing style, given minimal guidance. While we rely on random sampling for few-shot examples in our main experiments, we also conduct follow-up analyses exploring whether sampling based on topic or length similarity can further enhance stylistic similarity, among others.

\section{Experimental Setup}
In this section, we outline the key experimental choices of our study, with further details provided in Appendix~\ref{app:experimentalDetails}.




\paragraph{LLMs and Prompting} We consider both proprietary and open-weights models, including GPT-4o and GPT-4o-mini \cite{openai2024gpt4ocard}, Gemini-2.0-Flash \cite{google2024gemini2}, Gemma-3-27B \cite{gemmateam2025gemma3technicalreport}, DeepSeek-V3 \cite{deepseekai2025deepseekv3technicalreport}, and Llama-4-Maverick \cite{Llama-4}. The last two LLMs are state-of-the-art mixture-of-expert models, where DeepSeek-V3 has 671B total parameters with 37B activated for each token, and Llama-4-Maverick has 400B total parameters with 17B active parameters. We always prompt these LLMs with a 0 temperature to maximize output reproducibility. 

\paragraph{Writing Example Sampling} For each test sample, we provide 5 writing examples randomly sampled from the related train set written by the same author, shared across LLMs for stylistic mimicry. We employ random sampling since in principle we have no control over writing examples a user may include for LLM prompting. This ensures a more realistic evaluation.

\paragraph{Test Samples Summarization} We prompt GPT-4.1 \cite{gpt-4.1} to summarize each test sample.

\paragraph{Baseline} As a baseline, we prompt LLMs with the same test samples zero-shot (with only a content summary). As it is more or less expected that LLMs would perform worse in the style imitation task when not exposed to implicit personal writing style from provided writing examples than with ones. We use this simple baseline to (1) empirically validate this expectation and (2) to sanity check our evaluation framework proposed in Section~\ref{sec:methodology}. As a proof of concept, we only run four LLMs zero-shot, excluding DeepSeek and Llama models.

\paragraph{Evaluation Models} We train AA and AV models using transformer-based encoders, specifically Longformer-base-4096 and ModernBERT-base (see Appendix~\ref{app:experimentalDetails} for details). For each author, a style model based on LIWC \cite{boyd2022liwc} and WritePrint \cite{abbasi2008writeprints} feature sets is also constructed by extracting distributional stylistic features from their writing samples. We use these three models to evaluate the stylistic fidelity of both human-written (see right below) and LLM-generated texts (see the following sections). In addition, we employ GPTZero as the AI detection tool introduced in Section~\ref{sec:evaluation} off the shelf.

\paragraph{Validating the Evaluation Framework on Human Texts} To verify the reliability of our evaluation framework, we first report the evaluation results of the framework on original human-written texts (Table~\ref{tab:datasets}). This experiment reflects an ideal condition, where the four evaluators operate solely on human-authored data, allowing us to assess their upper-bound performance.

Across the four datasets (Enron, Blog, CCAT50, and Reddit), the AV models achieve high accuracy, ranging from 87.7\% to 91.4\%, and the AA models also perform decently, with top-5 accuracy ranging from 79.8\% to 95.5\%. These results suggest that both models are well fine-tuned and sufficiently reliable for evaluating authorship consistency. In addition, the style model\footnote{Additionally, an XGBoost classifier trained on these feature representations achieves substantial top-1 authorship attribution accuracy: 46.16\% (CCAT50), 43.12\% (Enron), 36.04\% (Reddit), and 34.11\% (Blog), despite the high number of authors (50–150). These results demonstrate the strength of our stylometric framework in capturing individual writing styles.} shows consistent alignment between test texts and their corresponding author’s style. The matching accuracy ranges from 69.1\% to 81.9\%, which indicates how often test samples of an author is closer to that particular author’s style model compared to other authors’ style models. Finally, human-writen test samples are barely detected as AI-generated by GPTZero. 

Taken together, these results confirm that human-authored texts exhibit personalized and distinguishable stylistic patterns, justifying the effectiveness of our evaluation framework.

\section{Results\label{sec:results}}  

We present a comprehensive analysis of how effectively LLMs mimic personalized style under our evaluation framework. We also report semantic similarity with the original texts to confirm that generated outputs follow the given content while preserving stylistic cues from the examples.

\subsection{Authorship Attribution Accuracy}
Table \ref{table_AA_result} reports the average top-5 AA accuracy across datasets and models under 5-shot (default) and 0-shot (baseline) settings. Overall, few-shot prompting consistently outperforms zero-shot, confirming that providing prior samples helps LLMs generate text more stylistically aligned with the target author. Interestingly, even zero-shot generations achieve non-trivial top-5 AA accuracy, largely driven by the topical/content overlap of the writings of that individual. Since authorship depends on both content and style \citep{sari2018topic}, our AA models leverage both dimensions, which explains the above-random performance without examples. This underscores the need for other assessments (as we provide in this study) that can provide a holistic evaluation for personalized writing.

\begin{table}[]
    \centering
    \resizebox{\columnwidth}{!}{
    \footnotesize
\begin{tabular}{llcccc}
\toprule
\textbf{Model} & \textbf{Setting} & \textbf{CCAT50} & \textbf{Enron} & \textbf{Reddit} & \textbf{Blog} \\
\midrule
\textbf{GPT-4o}           & 5-shot & 86.64 & 59.65 & 27.12 & 39.39 \\
        & 0-shot & 83.88 & 28.62 & 18.77 & 17.22 \\
\textbf{GPT-4o-mini}      & 5-shot & 87.36 & 56.45 & 26.30 & 38.47 \\
    & 0-shot & 85.84 & 27.56 & 18.95 & 16.10 \\
\textbf{Gemini-2.0-Flash} & 5-shot & 92.17 & 59.56 & 35.59 & 44.34 \\
 & 0-shot & 87.72 & 28.93 & 19.14 & 19.62 \\
\textbf{Gemma-3-27B}      & 5-shot & 93.30 & 62.38 & 30.53 & 39.38 \\
     & 0-shot & 86.26 & 27.12 & 19.20 & 18.89 \\
\bottomrule
\end{tabular}
}
    \caption{Average top-5 AA accuracy across LLMs under few-shot (5-shot) and zero-shot prompting (higher score is better). For each generated text $t'$ and its reference text $t$, accuracy is scored 1 if the true author $a$ appears in the AA model’s top-5 predictions. Scores are averaged across two AA models, with each cell showing the mean accuracy over all test samples for the corresponding dataset, model, and setting.}
    \label{table_AA_result}
\end{table}

\begin{figure}[h]
    \centering
    \includegraphics[width=\linewidth]{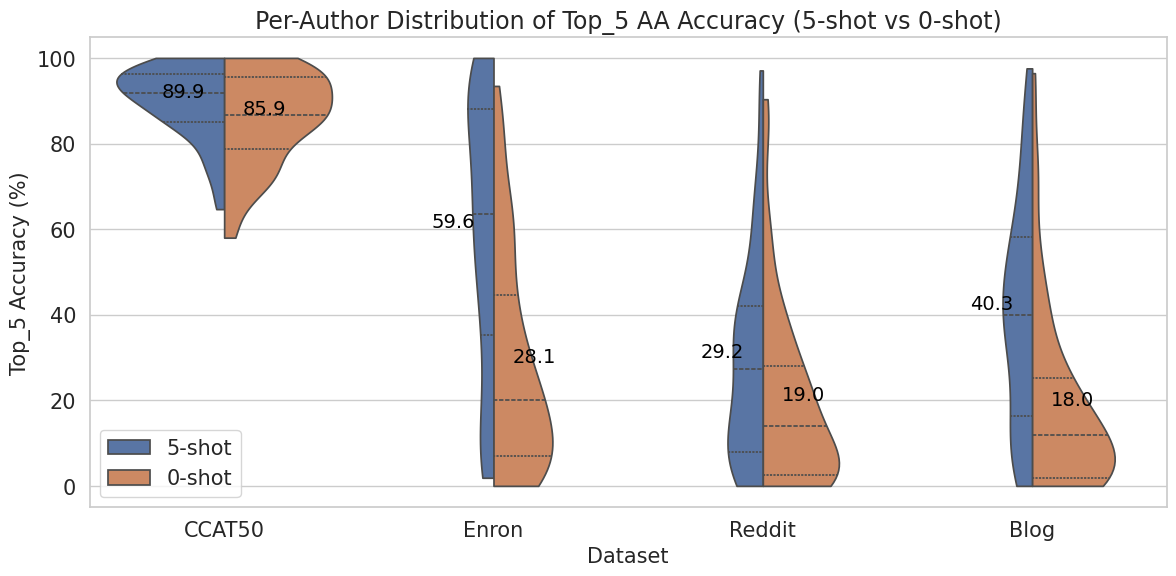}
    \caption{Distribution of per-author AA accuracy averaged across all LLMs under different settings. Highlighted values indicate the mean of each distribution. Overall, few-shot prompting achieves higher per-author accuracy than zero-shot.}
    \label{fig_AA_result}
\end{figure}

Accuracy also varies across individual users.
(see Figure \ref{fig_AA_result}). While CCAT50 yields consistently high accuracy, likely because journalists cover distinct topics, other datasets show greater variability. We observe higher accuracy in few-shot prompting than in zero-shot consistently for all authors. Notably, Gemini-2.0-Flash and Gemma-3-27B perform comparatively well on informal domains (see Table~\ref{table_AA_result}), such as Reddit and blogs, where stylistic cues include characters (*, \#, !) that vary widely across authors. Overall, these findings highlight both dataset-specific and model-specific sensitivities in personalized style replication.

\subsection{Authorship Verification Accuracy}
\begin{table*}[t]
    \centering
    \small

\begin{tabular}{llcccccc}
\toprule
\textbf{Dataset} & \textbf{Setting} & \textbf{GPT-4o} & \textbf{GPT-4o-mini }& \textbf{Gemini-2.0-Flash} & \textbf{Gemma-3-27B} & \textbf{DeepSeek-V3} & \textbf{Llama-4-Maverick} \\
\midrule

\textbf{CCAT50} & 5-shot & 95.28 & 95.16 & 97.44 & 97.34 & 97.46 & 94.68 \\
 & 0-shot & 93.14 & 93.38 & 93.94 & 94.18 & - & - \\
\textbf{Enron} & 5-shot & 96.15 & 96.64 & 96.44 & 96.17 & 96.30 & 95.65 \\
 & 0-shot & 85.02 & 85.68 & 85.95 & 84.98 & - & - \\
\textbf{Reddit} & 5-shot & 63.65 & 60.23 & 65.88 & 55.54 & 49.97 & 65.10 \\
 & 0-shot & 56.66 & 54.28 & 53.25 & 48.75 & - & - \\
 \textbf{Blog} & 5-shot & 19.37 & 17.93 & 21.25 & 16.72 & 20.77 & 17.61 \\
 & 0-shot & 8.15 & 8.22 & 10.35 & 9.55 & - & - \\
\bottomrule
\end{tabular}

    \caption{Average AV accuracy for various LLMs prompted under 5-shot and 0-shot settings, averaging over two AV models. Here, we assume that LLM-generated samples and the corresponding test set samples are from the same authors, so the accuracy is the higher the better.}

    \label{tab:AV_main}
\end{table*}

Table~\ref{tab:AV_main} presents (top-1) AV accuracy results for LLM-generated texts across the four datasets.  Across all models and datasets, the 5-shot setting consistently outperforms the 0-shot condition in AV accuracy. This shows that providing even a few writing examples significantly improves an LLM's ability to imitate implicit personal writing style. Notably, we also observe variation in AV performance across datasets like the AA results. Models perform particularly well on CCAT50 and Enron, which feature more structured and formal writing. In contrast, performance is generally lower on Reddit and Blog, where writing tends to be more informal and stylistically diverse.

While most LLMs, including Gemini-2.0-Flash, Gemma-3-27B, and DeepSeek-V3, show similarly strong AV performance across datasets, we observe that their effectiveness in mimicking authorial style still depends heavily on prompt design and the stylistic nature of the dataset. These results highlight the sensitivity of LLM-based authorship imitation to both task setup and domain characteristics.

\subsection{Stylistic Modeling and Alignment}

We build individualized style models adopting feature sets from LIWC \cite{boyd2022liwc}, which integrates linguistic and psychological cues, and WritePrints \cite{abbasi2008writeprints}, which contains syntactic-lexical patterns. To assess stylistic alignment, we compute the Mahalanobis distance \cite{mclachlan1999mahalanobis} between each test sample, whether human-written or LLM-generated, and the corresponding author’s style model.

\begin{figure}[t]
    \centering
    \begin{subfigure}{0.24\textwidth}
        \centering
        \includegraphics[width=\linewidth]{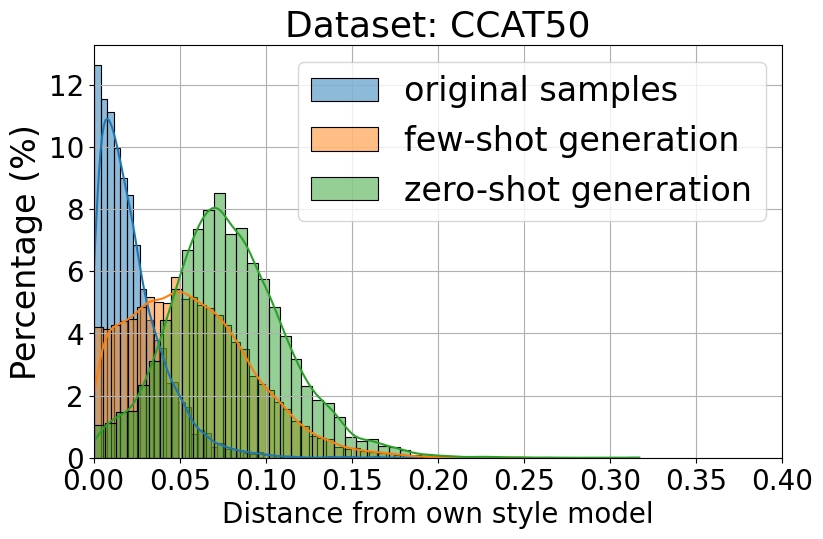}
      
    \end{subfigure}%
    \begin{subfigure}{0.24\textwidth}
        \centering
        \includegraphics[width=\linewidth]{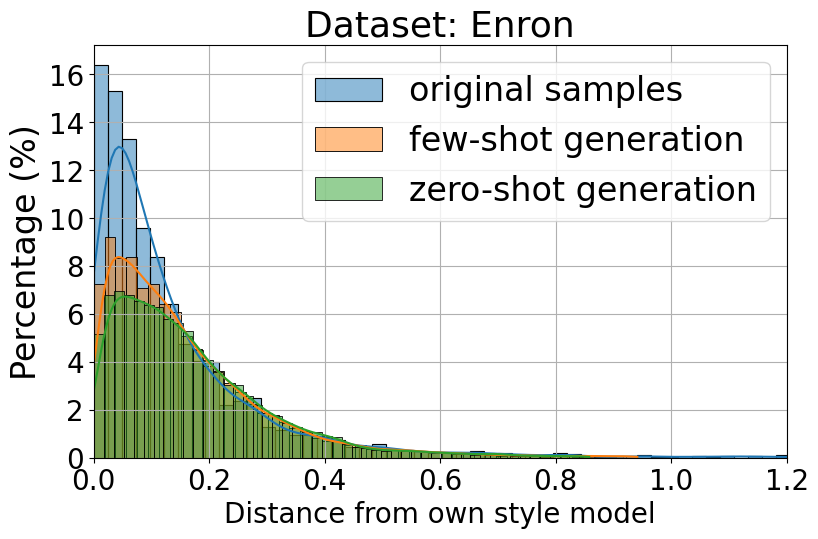}
      
    \end{subfigure}

    \vspace{0.25cm}

     \begin{subfigure}{0.24\textwidth}
        \centering
        \includegraphics[width=\linewidth]{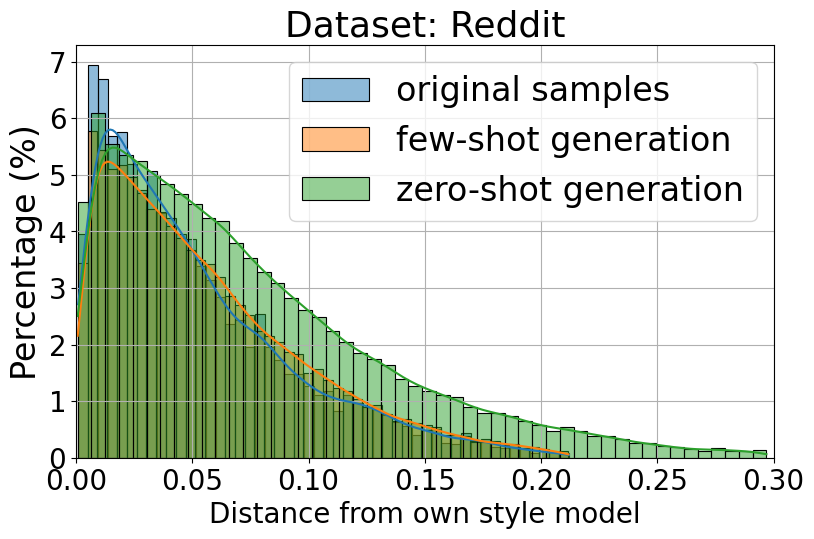}
      
    \end{subfigure}%
    \begin{subfigure}{0.24\textwidth}
        \centering
        \includegraphics[width=\linewidth]{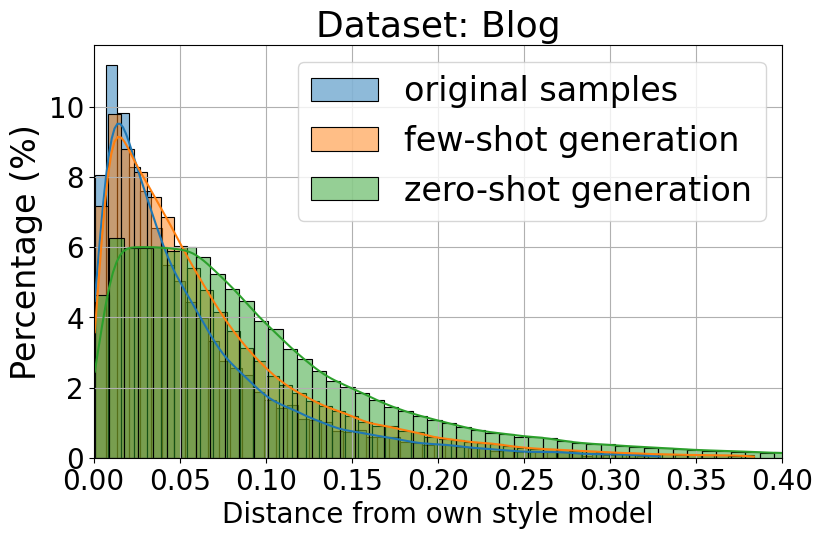}  
      
    \end{subfigure}
    
    \caption{Distribution of the average Mahalanobis distances between each text and the related target author's style model. 
Lower distances indicate more explicit stylistic similarity with the target style model. }
    \label{fig_style_model_results}
\end{figure}

Figure \ref{fig_style_model_results} illustrates the distribution of average distances between each text and the corresponding author’s style model across over all the test samples of each author under both 5-shot and 0-shot settings. As expected, original human samples exhibit the lowest distances. Crucially, few-shot generations are significantly closer to the target style than zero-shot ones, as confirmed by a Wilcoxon signed-rank test \cite{conover1999wilscon_ttest}. It aligns with our hypothesis that without example prompts, LLMs default to a generic style. We also observe variation across datasets, consistent with trends in authorship attribution and verification performance.

\subsection{AI Generation Detection}

Table~\ref{tab:AI_detect_human} reports the percentage of generations identified as human by GPTZero. This serves as a proxy for perceived naturalness and stylistic authenticity.

Overall, the percentage of LLM-generated texts classified as human-written is relatively low across most models and settings (i.e., always below 55\% and often less than 20\%). This means that while LLMs may generate texts with characteristics of an author, fooling specialized AA and AV models to varying extents, successfully passing as human-like generations remains a challenging task for LLMs.

Interestingly, GPT-4o and GPT-4o-mini show near-zero detection rates. In contrast, Gemini-2.0-Flash and Gemma-3-27B achieve significantly higher detection rates, especially on more structured datasets like Enron and CCAT50. We hypothesize that GPTZero may be optimized for detecting outputs from GPT-based models. As such, differences in detection rates should be interpreted cautiously, as they may reflect detector bias rather than true stylistic similarity. In line with previous sections, zero-shot setting consistently results in lower human-like detection rates across most models and datasets. Moreover, factors such as prompt design, model architecture, and the characteristics of the target domain, may play a role in the observed detection rates.

\subsection{Semantic Similarity} 

In our experiments, we provide a content summary in the prompt to guide the semantics of LLM-generated text, so that we can focus on writing style in our evaluation. As a sanity check, we measure the semantic similarity between the LLM-generated texts and the related human texts to see if LLMs do follow instruction to output a semantics-compliant generation. Table~\ref{tab:semantic_similarity} shows the similarity scores using METEOR, ROUGE \cite{kumar2024longlamp} and SBERT \cite{sbert}.

The scores indicate that LLM-generated texts are semantically similar to human-written ones (with SBERT scores $\geq$ 0.74) , demonstrating their strong instruction-following capabilities. Additionally, 5-shot prompting consistently produces slightly higher semantic similarity scores across all three metrics, aligning with the improved stylistic fidelity observed in our previous analyses. We hypothesize that higher stylistic fidelity, reflected in elements such as word choice and phrasing, may contribute to higher elevated semantic similarity scores.

\begin{table}[]
   \centering
    \resizebox{\columnwidth}{!}{
   \begin{tabular}{llrrrr}
\toprule

\textbf{Model} & \textbf{Setting} & \textbf{CCAT50} & \textbf{Enron} & \textbf{Reddit} & \textbf{Blog} \\
\midrule
{\textbf{GPT-4o}}           & 5-shot  & 0.08  & 16.86 & 0.67  & 0.47 \\
         & 0-shot  & 0.16  & 0.10  & 0.27  & 0.07   \\
{\textbf{GPT-4o-mini}}     & 5-shot  & 0.00  & 13.76 & 0.12  & 0.35 \\
     & 0-shot  & 0.08  & 0.08  & 0.19  & 0.12   \\
{\textbf{Gemini-2.0-Flash}} & 5-shot  & 20.49 & 54.25 & 20.77 & 18.74 \\
 & 0-shot  & 0.44  & 1.88  & 1.18  & 0.79 \\
{\textbf{Gemma-3-27B}}      & 5-shot  & 31.36 & 43.63 & 16.09 & 7.58 \\
     & 0-shot  & 0.20  & 2.65  & 4.80  & 1.67 \\
\bottomrule
\end{tabular}
} %
    \caption{\% of LLM generated texts detected as human text. The higher the number, the better.}
    \label{tab:AI_detect_human}
\end{table}

\begin{table}[]
\centering

\resizebox{\columnwidth}{!}{

\begin{tabular}{llccc}
\toprule
\textbf{Model} & \textbf{Setting} & \textbf{METEOR} & \textbf{ROUGE-L} & \textbf{SBERT} \\
\midrule
\textbf{GPT-4o }    & 5-shot & 0.32 & 0.24 & 0.77 \\
                       & 0-shot & 0.30 & 0.21 & 0.76 \\
\textbf{GPT-4o-mini} & 5-shot & 0.31 & 0.23 & 0.77 \\
                       & 0-shot & 0.29 & 0.20 & 0.76 \\
\textbf{Gemini-2.0-Flash}       & 5-shot & 0.32 & 0.26 & 0.78 \\
                       & 0-shot & 0.31 & 0.21 & 0.75 \\
\textbf{Gemma-3-27B}         & 5-shot & 0.30 & 0.23 & 0.75 \\
                       & 0-shot & 0.29 & 0.19 & 0.74 \\
\bottomrule
\end{tabular}
}

\caption{Average content similarity scores between LLM-generated and human reference texts.}
\label{tab:semantic_similarity}

\end{table}

\section{Follow-Up Studies\label{sec:analysis}}

Everyday users may go beyond providing random examples and a content summary to achieve more personalized writing. Rather than relying on random few-shot examples, we hypothesize that carefully chosen samples can better facilitate in-context learning \citep{kapuriya2025exploring}. Since authorship depends on both content and stylistic factors that vary with text length \citep{tripto2025beyond}, we select examples based on their content similarity and comparable length to the target text. Additionally, we provide the initial portion of the original text to guide generation \citep{bhandarkar2024emulating} and prompt with varying  number of examples. To examine these effects, we  evaluate on carefully sampled subsets of the test sets using GPT-4o, Gemini-2.0-Flash, and LLaMA-4-Maverick, ensuring cost efficiency and an apples-to-apples comparison. The following subsections present the ablation setups and key findings.

\subsection{Conditions Considered}

\paragraph{Content Similarity } (\texttt{+Sim ctrl}) We employ BERTopic \cite{bertopic} to cluster each author's writing, using both train and test samples. For each test sample, we select 5 train set samples that belong to the same cluster as the test sample. We note that this approach that selects few-shot exemplars based on semantic similarity operates in a manner similar to RAG \cite{mysore2024pearl}.

\paragraph{Length Alignment} (\texttt{+Len ctrl})  For each test sample, we choose 5 train set samples whose lengths are closest to that test sample.

\paragraph{Exemplar Quantity} We consider different numbers of writing examples while prompting: 2, 4, 6, 8, and 10. To ensure comparability, each smaller set is a strict subset of the next larger set.

\paragraph{Snippet Inclusion} (\texttt{+Snippet}) Following \citet{bhandarkar2024emulating}, we augment the prompt with an initial excerpt from the target test sample---the first 50 words or 20\% of the text (whichever is shorter)---besides the default 5-shot examples.

\begin{table}[t]
    \centering
\resizebox{\columnwidth}{!}{
\begin{tabular}{llcccc}
\toprule
\textbf{Dataset} & \textbf{Setting} & \textbf{AV} & \textbf{AA (top-5)} & \textbf{Style Acc.} & \% \textbf{Human} \\
\midrule
\textbf{CCAT50} & 5-shot & 96.39 & 89.72 & 60.11 & 8.50 \\
  & + Len ctrl & 96.83 & 90.16 & 57.89 & 5.67 \\
  & + Sim ctrl & 91.95 & 81.05 & 61.00 & 10.33 \\
  & + Snippet$^\dagger$ & 95.45 & 87.72 & 64.44 & 15.50 \\ 
  \midrule
  
\textbf{Enron} & 5-shot & 95.44 & 69.33 & 72.11 & 36.83 \\
  & + Len ctrl & 95.89 & 71.44 & 61.22 & 23.67 \\
  & + Sim ctrl & 81.28 & 36.00 & 70.00 & 39.67 \\
  & + Snippet$^\dagger$ & 90.56 & 60.50 & 68.44 & 46.50 \\ 
  \midrule
  
\textbf{Reddit} & 5-shot & 68.07 & 35.43 & 71.80 & 10.90 \\
  & + Len ctrl & 70.07 & 35.60 & 56.60 & 51.50 \\
  & + Sim ctrl & 53.10 & 16.63 & 69.33 & 12.60 \\
  & + Snippet$^\dagger$ & 72.83 & 36.83 & 68.27 & 22.10 \\ 
  \midrule
  
\textbf{Blog} & 5-shot & 19.40 & 43.93 & 82.67 & 9.00 \\
  & + Len ctrl & 20.68 & 43.03 & 68.60 & 27.50 \\
  & + Sim ctrl & 10.33 & 22.13 & 82.65 & 8.60 \\
  & + Snippet$^\dagger$ & 15.04 & 38.69 & 80.39 & 21.70 \\
\bottomrule
\end{tabular}
}
    \caption{Follow-up results based on the same subsets of test sets used in Section~\ref{sec:analysis}. The reported results are averaging over the three LLMs. The \texttt{+ Snippet}$^\dagger$ results are evaluated with only the LLM continued generations, not including the initially provided author snippets.}
    \label{tab:follow-up-results} 
\end{table}

\subsection{Major Observations} 

Table \ref{tab:follow-up-results} presents our evaluation framework results averaged over the three LLMs under various prompt configurations. The baseline is the 5-shot setting, in which five writing examples are randomly selected from the same author. 

Content-based exemplar selection (+Sim ctrl) surprisingly reduces attribution performance, especially in Enron, Reddit, and Blog. While topical alignment improves, restricting exemplars to a narrow cluster appears to diminish stylistic diversity, making it harder for models to capture author-specific cues.

Length alignment (+Len ctrl) yields modest gains in attribution (notably for CCAT50 and Reddit), but it lowers style-model accuracy. Matching exemplar length seems to benefit surface-level consistency captured by AA/AV, yet inadvertently reduces variation that signals stylistic fidelity.

Snippet inclusion (+Snippet) yields the strongest boost in perceived human-likeness across datasets, with detectors more often classifying outputs as human-written. This highlights the power of seeding generations with authentic text fragments, even if attribution metrics show mixed results.

\begin{figure}
    \centering
    \includegraphics[width=\linewidth]{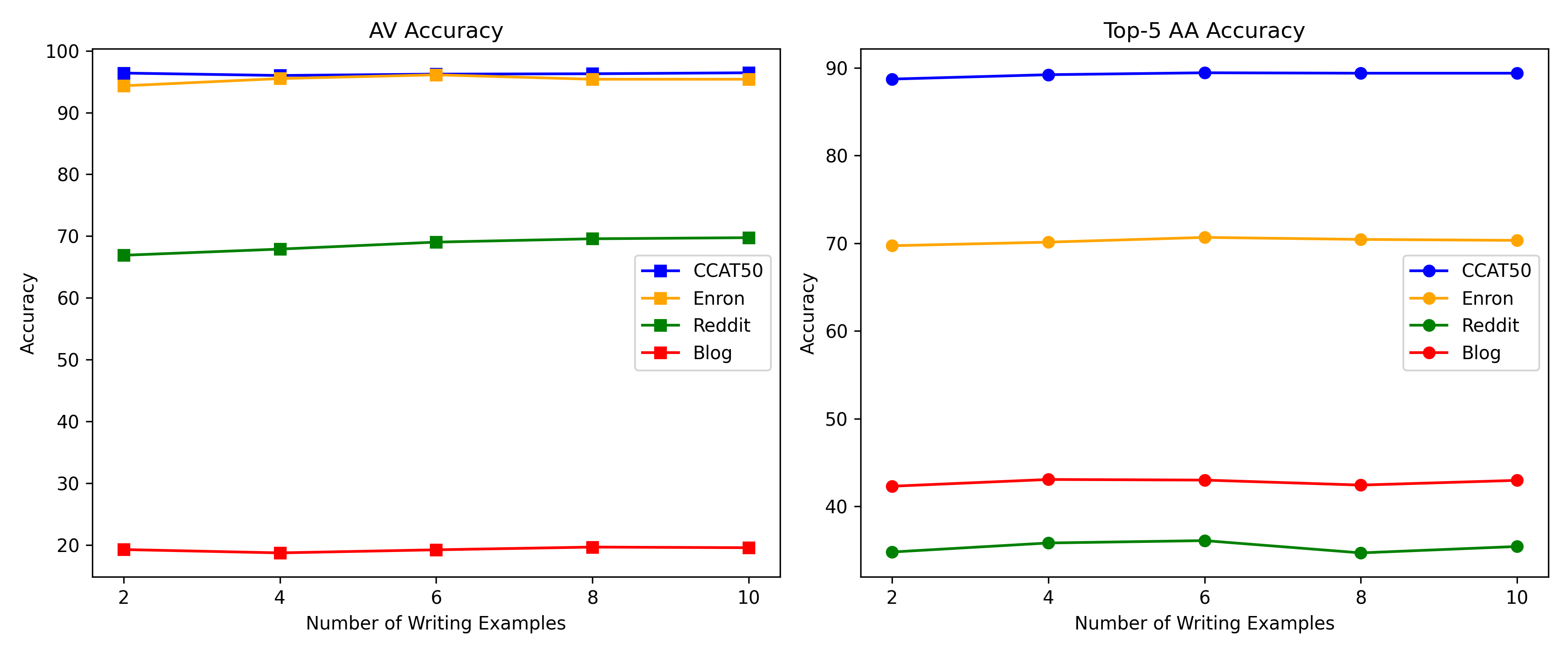}
    \caption{AA/AV accuracy as a function of the number of writing examples.}
    \label{fig:length_vs_accuracy}
\end{figure}

Interestingly, Figure~\ref{fig:length_vs_accuracy} shows that including more writing examples in the prompt affects the four metrics very little, suggesting limited gains in stylistic alignment. This is in line with the style model and AI detection results, which we include in Figure~\ref{fig:exemplar-quantity-style-ai-detection} in Appendix~\ref{app:follow-up} for space reasons.

Overall, these findings underscore that exemplar selection is far from trivial: strategies optimized for content or length do not always enhance stylistic imitation, and no single configuration consistently excels across all metrics. Effective personalization thus requires domain-sensitive choices and careful balancing between semantic control, stylistic fidelity, and human-likeness.

\section{Conclusion}

This paper presents a comprehensive evaluation of state-of-the-art LLMs on their ability to mimic the implicit writing styles of everyday users through few-shot in-context learning. By combining authorship attribution, verification, stylometric modeling, and AI generation detection, across four diverse datasets, we provide strong empirical evidence that, despite improvements from exemplar-based prompting, current LLMs still struggle to reproduce nuanced personal styles---especially in informal and stylistically diverse domains. Our analysis further shows that prompt design choices, such as length alignment and content similarity, moderately affect stylistic fidelity, but do not close the personalization gap. These findings highlight fundamental limitations in the stylistic adaptability of LLMs and suggest that achieving truly personalized generation remains an open challenge. Future work should explore richer personalization signals and hybrid prompting and/or finetuning strategies to better capture the subtleties of individual writing styles in real-world settings.

\section*{Acknowledgments}

Zhengxiang Wang was supported by Junior Researcher Award from the Institute for Advanced Computational Science (IACS) at Stony Brook University for the academic year of 2025. IACS also provided free OpenAI GPT access for our experiments. 
We thank the Google Gemma Academic Program for their partial support of Jiawei Zhou and for providing computational resources.


\section*{Limitations} 

Our study presents several limitations. First, while we propose a comprehensive computational evaluation framework, we do not include large-scale human evaluations. Manual assessments may reveal aspects of stylistic fidelity or perceived authorship not captured by automated metrics. 

Second, our analysis is constrained to within-genre personalization; that is, the writing genre of the prompt examples matches that of the generated text. Cross-genre style transfer remains unexplored due to the lack of multi-genre corpora authored by the same individual. 

Third, the generality of our findings is bounded by the four English-language datasets used, which, although diverse in domain, may not capture the full variability of real-world writing. 

Finally, our authorship analysis models are trained on relatively small sample sizes per author, which could limit their robustness, especially for authors with minimal stylistic consistency. 

Future work should incorporate broader linguistic and demographic diversity, include human evaluations, and explore personalization across modalities and genres.

\section*{Ethical Considerations}

\paragraph{Misuse Potential} Techniques for mimicking writing style may be misused for impersonation, academic dishonesty, or phishing. While our work aims to support personalized assistance, it also highlights risks associated with stylistic imitation.

\paragraph{Privacy} We use publicly available, anonymized datasets, but writing style can still carry identifiable traits. Care should be taken when using or releasing data for personalized modeling tasks.

\paragraph{Bias and Generalization} Our datasets are English-only and limited in demographic diversity. As a result, findings may not generalize to broader populations or multilingual settings, raising fairness concerns.

\paragraph{Evaluation Assumptions} We rely on computational authorship models, which may not fully reflect human perceptions of style and can encode unintended biases. These tools should be used cautiously in sensitive applications.

\paragraph{Disclosure and Transparency} As LLM outputs increasingly resemble human writing, clear disclosure of AI assistance is essential to maintain transparency and trust.

\bibliography{custom}

\clearpage

\appendix

\section{Train and Test split Preparation for Original Study}
\label{appendix_data_split}

As described in Section \ref{subsec_dataset}, we maintain separate train and test portions for each dataset. The train portion is used to train the AA and AV models, build style models for individual authors, and sample examples for in-context learning during LLM prompting. The test portion is reserved for evaluating AA, AV, and style models, as well as for conducting our main study (few-shot and zero-shot experiments).

For \textbf{CCAT50 (Reuters)}, we adopt the original train/test split from the source paper \cite{liu2011reuter}.

For the \textbf{Blog} dataset \citep{schler2006blog_corpus}, due to its large size and variability in text length, we restrict samples to 100–1500 words and select the top 100 authors by sample count. For each author, 50\% of samples are used for training and 50\% for testing, ensuring comparable distributions across length, topic, age, and gender demographics.

For the Enron dataset, we follow the preprocessing steps outlined in \citep{tripto2025beyond}, which include retaining only one-to-one internal emails to preserve a personal tone, excluding automated/forwarded/bulk messages, and removing emails with attachments. We further filter by length (100–1500 words), select the top 150 authors, and split the samples evenly into training and testing sets.

For \textbf{Reddit} dataset, where style can vary widely across subreddits \citep{volske-etal-2017-reddit_dataset}, we apply similar filtering (100–1500 words, top 100 authors by post count). To mitigate subreddit-specific bias, we stratify the train/test split such that each author’s samples maintain approximately equal subreddit distributions.

This consistent preprocessing ensures fair comparisons across datasets while balancing sample size, author coverage, and stylistic variation.

\section{Experimental Details on AA and AV Model Training}
\label{app:experimentalDetails}

\paragraph{Hyperparameters} Table~\ref{tab:training_hyperparams} summarizes the training configurations used for the Authorship Verification (AV) and Authorship Attribution (AA) models. We trained both longformer-base-4096 \cite{longformer} and ModernBert-base \cite{modernBert} for these two types of models.

\paragraph{Datasets} For AA model training and evaluation, we used the train/test splits of the four datasets created for the purpose of evaluating LLM's writing style mimicking capabilities. For AV model training and evaluation, we construct label-balanced (pos:neg = 4:6) AV datasets by sampling text pairs from the original train/test splits of the four datasets. For training AV/AA models, we split 20\% of the train set for validation.

\begin{table}[]
\centering
\small

\resizebox{\columnwidth}{!}{
\begin{tabular}{@{}llcc@{}}
\toprule
\textbf{Category} & \textbf{Training Parameter} & \textbf{AV Model} & \textbf{AA Model} \\
\midrule
General & Number of training epochs & 10 & 20 \\
& Train batch size & 8 & 8 \\
& Eval batch size & 16 & 16 \\
& Max sequence length & 2048 & 2048 \\
\midrule
Optimization & Learning rate & 2e-5 & 2e-5 \\
& Weight decay & 0.01 & 0.01 \\
& Warmup steps & 500 & 500 \\
& Gradient accumulation steps & 4 & 4 \\
\midrule

Evaluation & Evaluation strategy & epoch & epoch \\
& Early stopping patience & 3 & 3 \\
& Load best model at end & True & True \\
& Metric for best model & F1 & Eval loss \\
& Greater is better & True & False \\
\midrule

Precision & Mixed precision (fp16) & True & True \\

\bottomrule
\end{tabular}
}

\caption{Training hyperparameters for Authorship Verification (AV) and Authorship Attribution (AA) models.}
\label{tab:training_hyperparams}
\end{table}

\section{Elaborations on our Result Reporting}

Our primary goal is to present a large-scale, systematic evaluation. Given the scale of our study---over 40,000 generations per model across 400 authors---we used the average results to represent the expected performance of each LLM in implicit writing style imitation. While it would have been ideal to report the four metrics on the basis of each author, it is not practical to do so, given the large number of authors we have for experiments. Moreover, it is beyond the scope of the current study to examine the performance variations of LLMs in mimicking different individuals.  

That said, we did include per-author AA accuracy across different datasets in Figure~\ref{fig_AA_result} and paired statistical tests (i.e., Wilcoxon signed-rank test) for the style model results under different prompting conditions reported in Figure~\ref{fig_style_model_results}. This only not enriches our analysis, but also reconfirms the observations we made from the results averaging over authors from the four datasets. 

Lastly, to minimize generation variability and thus maximizes reproducibility, we used greedy decoding (temperature = 0) throughout all experiments.

\section{Follow-Up Studies\label{app:follow-up}}

\paragraph{Subset Sampling} To ensure apples-to-apples comparison, we use the same subsets of the four datasets across all the follow-up studies. For each dataset, we sampled 10 test samples for each author. The numbers of authors for CCAT50, Enron, Reddit, and Blog are 30, 30, 50, 50, respectively. We chose these numbers based on the BERTopic clustering results to make sure that each selected test sample had at least 5 train samples from the related author.

\begin{figure}[]
    \centering
    \begin{subfigure}{0.25\textwidth}
        \centering
        \includegraphics[width=\linewidth]{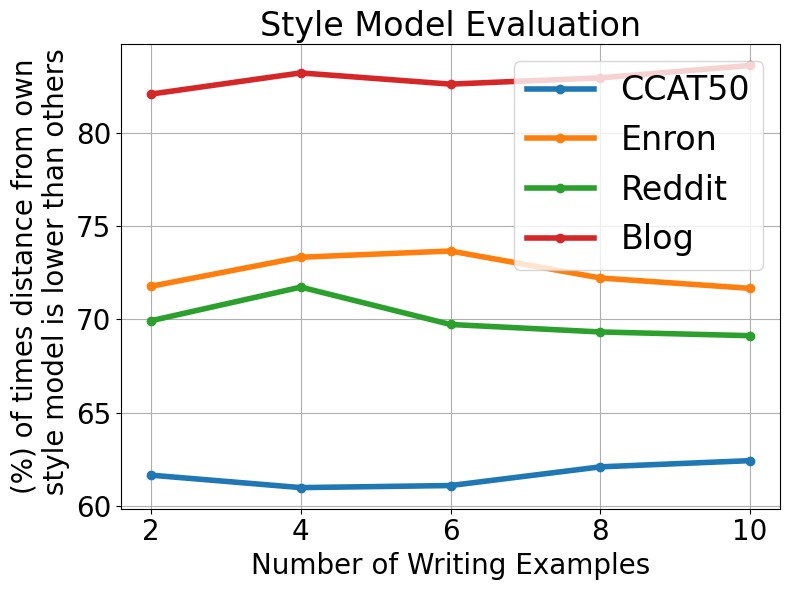}
    \end{subfigure}%
    \begin{subfigure}{0.25\textwidth}
        \centering
        \includegraphics[width=\linewidth]{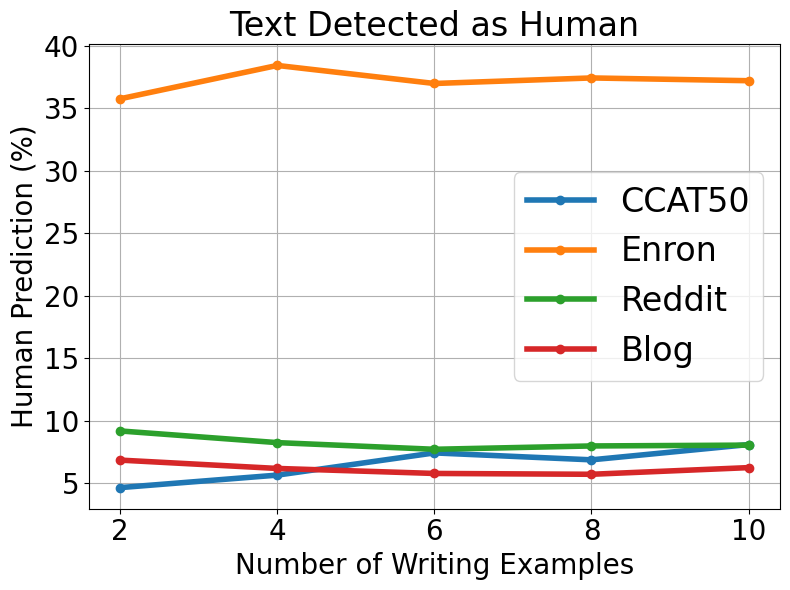}
    \end{subfigure}
    \caption{Style model accuracy and percentage of human detection rates as a function of the number of writing examples.}
    \label{fig:exemplar-quantity-style-ai-detection}
\end{figure}

\paragraph{Additional Result} Figure~\ref{fig:exemplar-quantity-style-ai-detection} shows the effect of including more writing examples in the prompt on the style model and AI detection results.

\section{Prompts\label{app:prompts}}

This section provides all the prompt templates used in this study. We use ``\$'' to denote placeholder.

\subsection{Prompt Template for Test Sample Summarization}

We used GPT-4.1 \cite{gpt-4.1} to summarize each sample in the test splits of the four datasets.

\begin{quote}
\small

You will be given a piece of text. Your task is to summarize the text in a concise and clear manner, capturing the main ideas and key points while maintaining the original meaning. \newline

\#\#\# Text to Summarize \newline

\$text \newline

\#\#\# Instructions \newline

- Provide a summary that is brief yet comprehensive.
- Ensure that the summary accurately reflects the content of the original text.
- Avoid adding any personal opinions or interpretations.
- Do not output anything other than the summary. \newline

Begin your response below:
    
\end{quote}

\subsection{Few-Shot Prompting for LLM Writing Generation \label{sec:few-shot-prompt}}

Besides 5-shot prompting in the main experiments in Section~\ref{sec:results}, we also re-use the following prompt templates in two of the follow-up studies in Section~\ref{sec:analysis}. They include Content Similarity  (\texttt{+Sim ctrl}) and Length Control  (\texttt{+Len ctrl}).

\begin{quote}
\small

You will be given one or more writing samples from a specific author. Your task is to analyze the author's style, tone, and voice, then craft a new piece of \$genre that closely mimics their writing based on a provided summary. Your writing should be around \$num\_words words. \newline

\#\#\# Author's Writing Sample(s) \newline

\$writing\_samples \newline

\#\#\# Writing Task Summary \newline

\$summary \newline

\#\#\# Instructions \newline

- Ensure your writing faithfully replicates the author's style, including tone, word choices, and sentence structure, etc.
- Maintain consistency with the author's voice while accurately reflecting the details of the given summary.
- Strive to make your writing indistinguishable from the original author's work.
- Do not output anything other than the writing. \newline

Begin your response below:
    
\end{quote}

\subsection{Zero-Shot Prompting for LLM Writing Generation}

We use zero-shot prompting only in the main experiments in Section~\ref{sec:results}, as a baseline to the default 5-shot prompting. 

\begin{quote}
\small

Given the following summary, your task is to generate a writing sample around \$num\_words words. The genre of the writing is \$genre. Do not output anything other than the writing.

\#\#\# Writing Task Summary

\$summary

Begin your response below:

\end{quote}

\subsection{Few-Shot Prompting Plus Snippet for LLM Writing Generation}

We include a text snippet, the first 50 words or 20\% of the text (whichever is shorter) into the default few-shot prompting illustrated in Appendix~\ref{sec:few-shot-prompt}.

\begin{quote}
\small

You will be given one or more writing samples from a specific author plus a text snippet of \$genre from the same author. Your task is to analyze the author's style, tone, and voice, then generate a continuation for the provided human-authored text snippet with around \$num\_words words that closely mimics their writing based on a provided summary. \newline

\#\#\# Author's Writing Sample(s) \newline

\$writing\_samples \newline

\#\#\# Writing Task Summary \newline

\$summary \newline

\#\#\# Human-Authored Text Snippet \newline

\$snippet \newline

\#\#\# Instructions \newline

- Ensure your writing faithfully replicates the author's style, including tone, word choices, and sentence structure, etc.
- Maintain consistency with the author's voice while accurately reflecting the details of the given summary.
- Strive to make your writing indistinguishable from the original author's work.
- Do not output anything other than the writing. \newline

Begin your response below:
    
\end{quote}

\end{document}